# Uncertain Inferences and Uncertain Conclusions

Henry E. Kyburg, Jr.
University of Rochester
kyburg@cs.rochester.edu

## Abstract

Uncertainty may be taken to characterize inferences, their conclusions, their premises or all three. Under some treatments of uncertainty, the inference itself is never characterized by uncertainty. We explore both the significance of uncertainty in the premises and in the conclusion of an argument that involves uncertainty. We argue that for uncertainty to characterize the conclusion of an inference is natural, but that there is an interplay between uncertainty in the premises and uncertainty in the procedure of argument itself. We show that it is possible in principle to incorporate all uncertainty in the premises, rendering uncertainty arguments deductively valid. But we then argue (1) that this does not reflect human argument, (2) that it is computationally costly, and (3) that the gain in simplicity obtained by allowing uncertainty in inference can sometimes outweigh the loss of flexibility it entails.

## 1   BEING UNCERTAIN AND BEING ABOUT UNCERTAINTY

There are many ways of expressing our uncertainty. In what follows, I shall try to ensure that nothing I say depends crucially on whether we express uncertainty by means of belief functions [Shafer, 1976; Smets and Kennes, 1994], or probability functions, or classes of probability functions [Levi, 1967; Levi, 1980], or probability envelopes [Walley, 1991], or by fuzziness [Zadeh, 1975], or by possibility measures [Dubois and Prade, 1985] , or any other way. However we construe uncertainty, though, we shall want to make a distinction between a sentence expressing that uncertainty, and the sentence whose uncertainty may be being expressed.

*Example 1*: "The probability of heads on a toss of a coin is $0.5 \pm \epsilon$." Barring the possibly question-begging occurrences of the indefinite article "a," this is simply a factual assertion about the behavior of coins [Reichenbach, 1949]. It embodies uncertainty, because one of the facts about coins is that the results of their flips are uncertain, and in fact uncertain in the way described. For most of us there is little or nothing uncertain about the statement itself.

*Example 2*: "Albert's degree of belief in the occurrence of heads on the next toss of this coin is 0.5." Again, this is a factual assertion about Albert; presumably about his willingness to buy and sell chances based on the outcome of the toss in question [Savage, 1954]. There is nothing special about Albert; we may in other contexts use the possessive pronouns "your," "our," and "my." A statement like this, in the context of reasoning or decision making, is a statement embodying uncertainty — perhaps in a different sense from that embodied in the first example.

Contrast those statements with these:

*Example 3*: "It is almost certain that the probability of heads on the toss of a coin is about a half." Several things may be observed. First, there is the obvious possibility that the interpretation of "almost certain" may be taken from a different representation than is given to probability; there is no reason why this cannot be interpreted subjectively while probability is interpreted in terms of frequencies. Second, the role in argument played by the quoted sentence in this example is quite different from the role played by the sentence in the first example.

*Example 4*: "It is very unlikely that Albert's degree of belief in the occurrence of heads on the next toss of this coin is 0.5." Similar observations may be made in this case. "Very unlikely" does not refer to Albert's degree of belief; it may be a value in some objective epistemic representation of uncertainty, or reflect a parameter in a statistical test, or represent the degree of belief of an experimental psychologist. As in the previous examples, the most important difference between Example 2 and Example 4 lies in the different roles they play in argument.

In order to say something useful about the roles these uses of uncertainty-language play in argument and inference, we must clarify the relation between argument and inference, and attempt to understand their role in our intellectual (and decision making!) economy.

## 2 ARGUMENT AND INFERENCE.

I am using "infer" as an action word: a person, or a machine, *infers* something from a set of premises or "makes an inference."[1] Like inference, reasoning is psychological. It is some thing that people do, and I see no reason why machines should not also be said to do it. I will take "argument," on the other hand, to be the formal counterpart of an inference or a piece of reasoning. Argument has an important social role — it helps us to come to agreement about what is the case: $A$ can use an argument to persuade $B$ that something is the case. Of course $A$ can use a lot of things other than argument — coercion, bribery, cajollery, sweet talk; and even among arguments some are better than others. Not all arguments are valid — $A$ can use argumenti ad hominum, ad populum, ..., ad infinitum.

Clearly, what we are interested in are *rational* arguments: arguments that are rationally compelling. Or which purport to be — I don't want to prejudge the possibility of defeasible arguments. I have avoided characterizing these arguments as "valid" both for the sake of allowing defeasible argumentation to be rationally persuasive, and for the sake of allowing the possibility of inductively good arguments.

A way of characterizing arguments that allows all these possibilities is as follows:

> An argument as structured in a formal language consists of
> 1. Premises.
> 2. A procedure of inference.
> 3. A conclusion

The premises of a good argument may be uncertain, or they may concern uncertainty, or they may involve both, or they may not involve uncertainty at all. The procedures of inference may be deductive, or inductive, or involve such inferential methods as hypothesis testing. The conclusion may be of categorical form, or may be qualified by a measure of uncertainty or may express uncertainty.

These possibilities can be captured with the help of a notation suggested by Hempel [Hempel, 1961]:

Schema I

$$\frac{Un\ ertainty\text{-}Premise_1}{P(Conclusion) = r}$$

This represents an argument in which there are two premises, one of which expresses uncertainty (probability, belief, ...), and in which the conclusion is a statement expressing uncertainty explicitly.[2]

*Example 5*:

$$\frac{The\ Probability\ of\ rain\ tomorrow\ is\ [0.2, 0.3]}{P(The\ ground\ will\ be\ wet\ tomorrow) = [0.2, 1.0]}$$

Observe that, following Hempel's intention, the argument here is strictly deductive. However we construe uncertainty—whatever semantic underpinning we give it—the premises of this inference can be true *only if* the conclusion is true. At least this seems to be so for all the alternatives we have considered above.

The second argument form isolated by Hempel is this:

Schema II

$$\frac{Uncertainty\text{-}Premise'_1}{Conclusion'} = p$$

This represents an argument in which the conclusion is categorical, unqualified, but in which the procedure of inference is not deductive. It is possible for the premises to be true, while the conclusion is false. The qualification, $p$, which occurs next to the double lines, is intended to characterize, not the conclusion, but the procedure of inference.[3]

*Example 6*:

$$\frac{The\ probability\ of\ rain\ is\ .95}{The\ picnic\ will\ be\ a\ disaster}\ .95$$

The intended interpretation of this argument structure is this: the first premise contains knowledge of uncertainty (in the example, the uncertainty that it will rain). The second premise contains background knowledge regarded as categorical, uncompromised by uncertainty, so far as the context of this argument goes. This much is parallel to the first form of argument. But there are two crucial differences.

---

[1] I will forbear to start a new style by saying that a person "inferences" a conclusion.

[2] The letter $P$ which is used to measure the uncertainty of the conclusion should not prejudicially be taken as the first letter of a possible appropriate measure.

[3] 0.95 represents the minimal cogency of the argument in the following example.

First, the procedure of inference is no longer assumed to be deductive. Inference, on this pattern, is not necessarily truth preserving. It could be inductive inference, what I have called 'probabilistic inference' in other places [Kyburg, 1988], statistical inference in the classical form developed by Neyman and Pearson [Lehman, 1959], and perhaps various other forms of inference such as inference by analogy. Recall that the "procedure of inference" may involve many steps; all that is intended by this model is that at least some of the steps may not be truth preserving.

Second, the conclusion of the inference is no longer a statement of uncertainty concerning another statement, but is simply that statement itself. It is a categorical conclusion. In the example, the conclusion is *not* that the picnic will *probably* be a disaster, but that it categorically *will* be a disaster. This is not to say that the inference is incorrigible. Further evidence (for example a more recent weather report), added to the premises we have taken into account, could lead to the opposite conclusion, or could undermine the degree of warrant we have for the conclusion. But that is another matter, concerning another argument. By regarding the conclusion as categorical, we are just saying that in the present context, with the present evidence, we are not taking any alternative seriously. We are not willing to gamble against the conclusion.

## 3    THE INFERENCE PROCEDURE

In both schemata we have allowed both 'certain' and 'uncertain' premises. Let us first observe that this is perfectly general. There is no problem in taking all the categorical or unqualified premises of an argument and representing them by a single conjunction. Combining the premises concerning uncertainty is another kettle of fish. Such a premise can always be thought of as providing a distribution—perhaps so simple as a two point distribution representing the probability of a single statement, or perhaps so complex as a many dimensional joint distribution of several continuously varying quantities. Each of these premises can be regarded as giving the *marginal* distribution of one or several random quantities. To combine the premises concerning uncertainty, then, is to replace these marginal distributions (or sets of marginal distributions, or, ...) by a single joint distribution.

There are a number of ways of doing this, depending on the uncertainty representation with which we start. If we represent uncertainty by a single probability distribution, then we must introduce principles— for example, indifference, or maximum entropy—that will allow us to arrive at a single distribution. These principles then become part of what we are taking as premises for the inference. If we represent uncertainty by a set of probability distributions, or by upper and lower probabilities, then we can take the marginal distributions as imposing *constraints* on the joint distribution; we can do this without making substantive assumptions.[4] In any event, what we come to is that the uncertainties that are germane to the argument can be captured in a single statement of uncertainty—what we have represented as *"Uncertainty-Premise."*

Argument schema I may now be taken to have this form: there is a single premise expressing our knowledge of uncertainty. In the Bayesian ideal form, it consists of a probability measure over possible worlds, in belief function form it consists of a belief function defined on a frame of discernment, under other representations of uncertainty, it may have other forms. The second premise consists of a conjunction of categorical, unqualified, evidential, statements. Our conclusion is, in Hempel's terms, a statement of probability. But just as we generalized the idea of a premise of uncertainty, so we can generalize the conclusion in the same way we generalized the premise of uncertainty, and we can take the conclusion to be an uncertainty distribution (or any of the specific uncertainty statements derivable therefrom).

This said, the procedure of inference leaps out at us. In the probabilistic framework, the procedure is nothing less than classical conditionalization.[5] For belief functions it may be "Dempster conditioning" [Shafer, 1976]. For other representations of probability, the procedure may be different again. The point is that each representation of uncertainty has its own "updating" procedures.

We now claim that the procedure of inference in this case is classical deductive inference, and that the arguments that are valid are valid in the classical sense that their conclusions are true in every model in which the premises are true. In short, we claim that there is nothing "uncertain" about the inferences represented in an argument fitting the first schema.

Consider the probability case. We begin with a dis-tribution of uncertainty, $Premise_1$, and a premise consisting of a conjunction of categorical statements, $Premise_2$. We derive $P(Conclusion) = r$ by conditionalizing our initial $Uncertainty\text{-}Premise$, $Premise_1$, on $Premise_2$.

---

[4]It sometimes suggested argued that the the premises can be characterized as "assumptions" thereby avoiding responsibility for their truth or correctness or suitability. We will not take this assumption seriously.

[5]With a little artificiality it is easy to get the effect of "Jeffrey conditionalization" in the classical framework [Kyburg, 1986].

*Example 7*: We are conducting a compound experiment: drawing an urn from a population of two types of urns, and drawing a single ball from the urn selected, and noting whether it is black or white. Our knowledge of uncertainty is represented by a joint distribution according to which the fraction of urns of type I is 0.6, and the chance of a black ball from an urn of type I is 0.9, and the chance of a black ball from an urn of type II is 0.2. (Equivalently: we select a patient from a pool of patients, of whom 60% carry virus $v$; we perform a test which gives a positive reaction 90% of the time in carriers, and only 20% of the time in non-carriers.) We add as a premise that ball $a$, the ball we have drawn, is black. (The test has given a positive reaction.) We conclude that the probability is 0.87 that the urn we have tested is of type I (that the patient is a carrier).

|       | type I | type II |
|-------|--------|---------|
| black | 0.90   | 0.20    |
| white | 0.10   | 0.80    |

This matrix represents the chance that the ball we have chosen, $a$, is black according to the two kinds of urns from which we might have selected it.

This is combined with the knowledge that the chance of getting type I is 0.60 and the chance of getting type II is 0.40, i.e., that the probability that the selected urn $u$ is of type I is 0.60 and that it is of type II is 0.40.

These two facts represent our knowledge of uncertainty.

Our categorical knowledge is represented by the statement that we have selected urn $u$, and that the ball we have drawn from urn $u$, ball $a$, is black.

Our conclusion (that the probability is 0.87 that the urn we have tested is of type I) is formally derivable from these two premises: we may formally deduce the sentence

$P(Urn\ u\ is\ of\ type\ I | ball\ a\ is\ black) = 0.87$

from the two premises.

There is a minor bit of legerdemain here; namely, one would ordinarily construe the table as representing a general bit of statistical knowledge, and we need to be careful in applying it to the specific ball $a$. Similarly, we are applying our general knowledge about the frequency of urns of the two kinds to $u$. But this is just the question of determining "the reference class" for the application of statistical knowledge; some of us do regard that as a serious problem, but it is not one I want to address here [Kyburg, 1983].

The example illustrates a fairly uncontroversial and simple bit of Bayesian inference. We examined it in some detail because we wanted to be very clear that the *inference* involved is pure, mathematical, deductive, truth-preserving inference.

While other forms of uncertainty representation and updating will employ different bits of mathematics, the general principle is the same. The measure of uncertainty employed will be defined in such a way as to make use of certain logical and mathematical machinery, and the application of that machinery in a specific case will be no more than the application of deductive procedures to that specific case.

## 4 WHAT ARE ARGUMENTS FOR?

Arguments no doubt serve a lot of functions, but one is to aid in decision making. For this purpose, arguments conforming to the first schema are exactly what we need. We need to assess the uncertainty of the various consequences of various actions in order to compute the expected utility of each act open to us. In the simplest case, the argument results in a number that can be construed as a probability. These probabilities are the probabilities that, combined with utilities, yield expectations. If we agree on utilities and agree on the evidence, we should agree on acceptable actions.

Another function of argument is to establish those statements we take as premises in arguments. For this function, the second schema seems right. We often want to accept a statement categorically, without hedging. This is most clear with respect to statements that are close to reports of experience—exactly those statements that, under the first schema, we would be disposed to conditionalize on.

*Example 8*: Suppose we have to decide whether or not to hold a picnic tomorrow. You and I agree, at least roughly, on the relative utilities of holding a picnic if it rains, calling the picnic off if it rains, holding the picnic if it doesn't rain, and calling the picnic off if it doesn't rain. One of the factors involved in assessing the chance of rain is the barometric pressure. We have accepted a general statistical theory of the weather (let us assume), and applied all the knowledge we have to arrive at a probability of rain conditional on the barometric pressure. To obtain an updated probability of rain, all we have to do is plug in the barometric pressure. But the theory we have does not give the probability of rain conditional on our having read a certain number; it gives that conditional probability in terms of the *actual* pressure. The actual pressure is what we need to plug in; but we can only obtain it by means of *inference* from the "reading." We must uses schema II to infer a premise we need for schema I.

There thus appears to be a need for both types of conclusion and thus both types of inference.[6]

## 5  ARE UNQUALIFIED CONCLUSIONS NECESSARY?

A natural question to ask is whether there is really a need for both types of inference, or whether the apparent need is illusory. There is a sense in which we might say that the need is illusory. Suppose that one item, say $P$, in the categorical premise $P_2$ needed for an inference of the first sort is justified by means of an argument fitting the second schema. As our gloss on the first schema showed, in a probability framework, we may represent the probability of the conclusion as the conditional probability of the conclusion *given* the categorical evidence $P_2$.

As our gloss on the second schema showed, this means that, according to the uncertainty distribution called upon by that argument, the uncertainty of $P$, updated by $P_2'$, is less than $1 - p$ (where $p$ is our acceptance level).

Let us suppose that the uncertainty premises of both inferences are consistent with each other, and can justifiably be combined. (This requires a lot more than bare consistency.) We can now construct a schema of the first form that does *not* require the categorical premise $P_2$.

The argument has as premises the combination $C(P_1, P_1')$ of the uncertainty premises of the two old arguments. As a categorical premise it uses the old categorical premise with the component $P$ deleted, but with $P_2'$ from the second argument added. The uncertainty of the conclusion is modified to take account of the uncertainty of $P$. We compute:

$P(Conclusion) =$
$P((Conclusion \wedge P) \vee (Conclusion \wedge \neg P))|P_2 \wedge P_2') =$

$P((Conclusion \wedge P)|P_2 \wedge P_2')+$
$\qquad P((Conclusion \wedge \neg P)|P_2 \wedge P_2') =$
$P(P|P_2 \wedge P_2')P(Conclusion|P \wedge P_2 \wedge P_2')+$
$\qquad P(\neg P|P_2 \wedge P_2')P(Conclusion|\neg P \wedge P_2 \wedge P_2')$

Since $P(P|Premise_2)$ is high and $P(\neg P|Premise_2)$ is low, this sum will be close to the old value, $P(Conclusion|P \wedge P_2')$

*Example 9*: To calculate the likelihood of rain, we conditionalized on the barometer *reading*, even though we knew that the barometer was not a perfect measuring instrument. If we are challenged on the basis of the fact that our barometric observation merely confers probability on our conclusion that the pressure has a certain value, we may note that (1) we can combine the uncertainty knowledge we have about the barometer with that concerning the general conditions of the picnic, (2) we may combine the barometer observation with the other categorical statements relative to which we are calculating the desirability of holding a picnic, and (3) we may compute the conditional probability of the picnic being a disaster relative to this new collection of evidence. We note that it is close to the original conditional probability.

Under what conditions will such a procedure work? We have demonstrated it for probability as the measure of uncertainty, but even in that case there are some conditions that must be satisfied. As we mentioned, not only must $Premise_1$ be consistent with $Premise_1'$, but it must be reasonable to adopt a joint distribution embodying both of them. In any situation in which an argument of schema II may be used to justify the use of a premise $P$ in an argument of schema I, it is not hard to see how this condition could be met, in terms of any representation of uncertainty, provided the original inferences were satisfactory.

The second condition is more problematic. In our discussion we depended on the fact that probability is additive, and that if the probability of $P$ is "high", then the probability of $\neg P$ is "low". Again, however, when "high" is construed as allowing for unqualified detachment of a conclusion from the premises that make it relatively certain, it is hard to see how this condition could fail to be met either. In terms of belief functions, for example, a "high" $Bel$ value for $P$ must be associated with a "low" $Bel$ value for $\neg P$, and $Bel$ applied to an exclusive disjunction must have at least as high a value as the sum of its values for the disjuncts [Shafer, 1976]. Similarly, possibility theory [Dubois and Prade, 1985] seems to satisfy this condition, as does the transferable belief model [Smets and Kennes, 1994].

---

[6]Attempts have been made [Levi, 1967; Levi, 1980] to unify these two functions. The idea is that we should regard the acceptance of a statement as an act that, like any other, has consequences that depend on the actual state of the world. The general idea goes back to Dewey [Dewey, 1938], and perhaps earlier. One difficulty with this idea is that it is hard to know what the long-run consequences of accepting a statement may be, when it is a statement that itself functions as a premise in an argument whose function is to take account of the more immediate consequences of particular actions.

## 6   THE ELIMINABILITY OF SCHEMA II

Since the use of schemata of the second form to generate categorical premises for schemata of the first form is, subject to the two conditions mentioned above, eliminable in the case of any particular categorical premise, we may ask whether it is eliminable in the case of the collection of categorical premises in schema I. Here an interesting possible problem arises. Let us suppose that the premises whose justification is sought via schemata of the second kind are $P_{21}, \ldots, P_{2k}$. What we need for updating the uncertainty distribution of the first argument is, in effect, the *conjunction* of $P_{21}, \ldots, P_{2k}$. In the context of schema I, where these are regarded as incorrigible and unqualified, there is no problem: we can conjoin certainties *ad libitum*. But when we reject these premises as certainties, and seek to incorporate the evidence for them into schema I, we must worry anew about lottery problems [Kyburg, 1961], just as we must in schema II when what we are looking for is an argument for the conjunction in question.

*Example 10*: We are considering inviting John to join our Frisbee team. Practise isn't important, but native athletic ability is. Our ignorance premise says that almost anyone who is under six feet tall and weighs more than 250 pounds is unathletic ($\neg A$). Our categorical premises say that John is under six feet tall, and weighs more than 250 pounds. Let us represent additional categorical premises by $E$. We conclude that the expected value of inviting him to join our team is less than the expected value of not inviting him to join our team.

But our categorical premises are not "certain"—they are supported by arguments of type II. Let the categorical premises of these arguments (which presumably include scale readings and stadiometer readings) be $F$, and suppose that they each support their conclusion (respectively, "John's weight is more than 200 pounds" and "John's height is less than six feet") just well enough to justify its use in a schema of type I.

Eliminating the use of schema II leads to the replacement of

$$P(\neg A(J)|E \wedge W(J) > 250 \wedge H(J) < 6),$$

which was high, by its development in terms of $W(J) > 250$ and $H(J) < 6$, conditioned on both $E$ and $F$. The result consists of four terms, the first of which contains the factor

$$P(H(J) > 250 \wedge H(J) < 6|E \wedge F)$$

and the remaining three contain, respectively, the factors

$$P(\neg A(J)|\neg W(J) > 250 \wedge H(J) < 6 \wedge E \wedge F)$$

$$P(\neg A(J)|W(J) > 250 \wedge \neg H(J) < 6 \wedge E \wedge F)$$

$$P(\neg A(J)|\neg W(J) > 250 \wedge \neg H(J) < 6 \wedge E \wedge F).$$

The last three factors could quite easily be low (lots of short people are athletic, lots of heavy people are athletic, and lots of people who are neither short nor heavy are athletic. Each of the four terms of the sum contains a factor that may be well below one, and their sum may well be less than degree to which schema II supports $\neg A(J)$

What we learn from this example might well be the advisability of eschewing schema II entirely. We cannot always replace its use in the way in which we have outlined, but the reason is exactly that its use is not always warranted. When it is not warranted is exactly when a number of arguments conforming to schema II are used to provide justification for the categorical premises of an application of schema I that requires conditioning on these premises simultaneously: To have good enough reasons for $A$ and for $B$ is *not* to have good enough reasons for $A \wedge B$. This is the lottery "paradox."

Carnap [Carnap, 1968], discussing the possibility of inductive logic using arguments conforming to schema II, argues that while people do seem to discuss arguments in the way suggested, it is to be understood as an abbreviation or shorthand for a more adequate view of induction, which should proceed strictly according to schema II. For Carnap there is no place for arguments conforming to schema II. This seems to be true of Cheeseman, as well [Cheeseman, 1988].

Nevertheless, there are some arguments in favor of schema II playing a role in uncertain inference.

## 7   ARGUMENTS IN FAVOR OF SCHEMA II

First, as many writers agree, if the use of schema II is only a shorthand approximation for the use of schema II, it is nevertheless an approximation that is useful and convenient. We may want to be careful about its use, but that is not grounds for eschewing it altogether.

Second, the fact that people do accept highly supported conclusions, rather than merely according them their appropriate degree of uncertainty, should not be dismissed lightly. I would be the last person to suggest that this reflects an intention on the part of the Grand Engineer, but it is certainly the case that our characteristics, including our epistemic characteristics, have been honed over many millenia. Much of what we do—not all, of course—*works*.

Third, consider what is involved in eschewing schema II. It is hard to know exactly what can go into *premise*$_2$ of that schema, but whatever it is, it must have the following characteristics: It must be consistent in the very strong sense that no conjunction of parts of *Premise*$_2$ can be unacceptable. It must be monotonic, in the sense that no amount of contrary evidence can led us to withdraw or modify any part of *Premise*$_2$. An extreme suggestion is to take perceptual judgments to fall in this category, but this can't be right: any *judgment*, including a perceptual judgment, can be wrong. Even these would not be monotonic in the required sense. It seems much more natural and reasonable to include as data judgments that we reasonably regard, not as incorrigible, but as practically certain. But of course *these* judgments are nonmonotonic. It is not clear that any statement having content can be regarded as incorrigible.

Fourth, if we are to include the evidential data supporting the wide class of propositions we need to take account of in our day to day lives, the sheer mass of the data could be overwhelming. Imagine how extensive the data would be for even quite straight forward arguments, and how complex the full arguments must be. For example consider the problem of reconstructing an engineering argument in terms of "perceptions" rather than in terms of quantities!

*Example 11*: Consider an argument of type I that involves conditioning on $k$ pieces of data, each of which is acceptable in terms of an argument fitting schema II, but none of which is incorrigible. Not only would we have to accommodate in *Premise*$_2$ all of the underlying immediate data that support *each* of the $k$ pieces of data, but measure of the conclusion will (if probability is a guide) consist of the sum of $2^k$ terms, each of which involves a complex operation of updating.

Fifth, we have so far attended only to the premises we have taken to be categorical. But in arguments of both kinds we need premises embodying our knowledge about uncertainty. According to some views, these premises neither have nor need justification: they are taken to represent the opinions of the agent. But if the "agent" is more than one person, and the persons involved do not agree on a prior distribution of probability, argument schema I, as a pattern that can be used for coming to agreement, breaks down.

According to other views, these premises represent the application of a principle such as a principle of indifference [Laplace, 1951] or a principle of maximizing entropy over a set of answers [Jaynes, 1957]. The justification for the premises is provided by these principles. While the former view seems too liberal to ensure agreement among cooperating agents, the latter seems too dictatorial.

There is a middle ground, of course, which is often encountered in the real world. This accepts the idea that knowledge of uncertainty is knowledge that has been earned by hard scientific inquiry, based on data that are objective and public. But if the uncertainty premises are taken to be *accepted* in arguments fitting schema I, their justification, since it is for their categorical acceptance, must be taken as conforming to schema II.

Several forms of nonmonotonic logic have been proposed; these may be construed as governing inferences conforming to schema II, though generally lacking in an uncertainty premise. Nevertheless, though leaving much about which agreement is needed implicit, they do yield categorical, non-qualified conclusions, which are uncertain in the specific sense that they may be withdrawn in the light of new evidence.

An attempt to provide a general semantics for inferences conforming to schema II has been made in [Kyburg, 1996]. Specifically statistical inferences have been explored in [Kyburg, 1974]: what is sought there is exactly that middle ground referred to above, in which our knowledge about uncertainty is, like other knowledge, based on objective and public data, and methods of inference that may be characterized either as yielding the truth frequently or as succumbing to error rarely.

## 8   CONCLUSION

Taking Hempel's paradigms to provide alternative frameworks for "uncertain" inference, we have seen that in principle it is possible to incorporate the uncertainty in the uncertainty premise of the inference, and that the uncertainty characterizing the conclusion is then given explicitly. This has the advantage of robustness, but the disadvantage of complexity. The inference itself, in such a schema, is not uncertain, but deductively valid. The (probabilistic) conclusion holds in every model in which the premises hold.

One disadvantage of such a schema is that it leads to a high degree of complexity in uncertain inference. It also fails to reflect the habits of human argumentation, which often employ schemata of type II to establish the premises needed for arguments falling under the first schema.

Schemata of type II have the disadvantage that they cannot be freely combined, since the justifications of two statements do not necessarily provide a justification for their conjunction. Further, the inferential pro-cedures they employ are not valid: there can be models of the premises in which the conclusion is false.

Uncertain inferences with categorical conclusions appear to be what have inspired those who work with nonmonotonic logic. They correspond to certain classes of classical statistical inference—confidence intervals, hypothesis testing [Lehman, 1959]. How to deal with the non-conjunctive character of such inferences is a problem that is under investigation in a number of areas, but it is one for which no general agreed-upon solution exists. Our considerations suggest that this is a worthy area to explore.

### Acknowledgments

Support for this work was provided by National Science Foundation grant IRI-9411267

# References

[Carnap, 1968] Rudolf Carnap. Inductive logic and inductive intuition. In Imre Lakatos, editor, *The Problem of Inductive Logic*, pages 258–267. New Holland, Amsterdam, 1968.

[Cheeseman, 1988] Peter Cheeseman. Inquiry into computer understanding. *Computational Intelligence*, 4:58–66, 1988.

[Dewey, 1938] John Dewey. *Logic: the Theory of Inquiry*. Henry Holt and Co., New York, 1938.

[Dubois and Prade, 1985] Didier Dubois and Henri Prade. *Possibility Theory: An approach to Computerized Processing of Uncertainty*. Plenum Press, New York, 1985.

[Hempel, 1961] Carl G. Hempel. Deductive-nomological vs statistical explanation. In Herbert Feigl, editor, *Minnesota Studies in the Philosophy of Science III*, pages 98–169. University of Minnesota Press, 1961.

[Jaynes, 1957] E. T. Jaynes. Probability theory in science and engineering. *Colloquium lectures in Pure and Applied Science*, pages 152–87, 1957.

[Kyburg, 1961] Henry E. Kyburg, Jr. *Probability and the Logic of Rational Belief*. Wesleyan University Press, Middletown, 1961.

[Kyburg, 1974] Henry E. Kyburg, Jr. *The Logical Foundations of Statistical Inference*. Reidel, Dordrecht, 1974.

[Kyburg, 1983] Henry E. Kyburg, Jr. The reference class. *Philosophy of Science*, 50:374–397, 1983.

[Kyburg, 1986] Henry E. Kyburg, Jr. Baysian and non-bayesin evidential updating. *Artificial Intelligence*, 31:271–293, 1986.

[Kyburg, 1988] Henry E. Kyburg, Jr. Probabilistic inference and probabilistic reasoning. In Shachter and Levitt, editors, *The Fourth Workshop on Uncertainty in Artificial Intelligence*, pages 221–228, 1988.

[Kyburg, 1996] Henry E. Kyburg, Jr. Combinatorial semantics. *Computational Intelligence*, 1996.

[Laplace, 1951] Pierre Simon Marquis De Laplace. *A Philosophical Essay on Probabilities*. Dover Publications, New York, 1951.

[Lehman, 1959] E. H. Lehman. *Testing Statistical Hypotheses*. John Wiley and Sons, New York, 1959.

[Levi, 1967] Isaac Levi. *Gambling with Truth*. Knopf, New York, 1967.

[Levi, 1980] Isaac Levi. *The Enterprise of Knowledge*. MIT Press, Cambridge, 1980.

[Reichenbach, 1949] Hans Reichenbach. *The Theory of Probability*. University of Califormia Prss, Berkeley and Los Angeles, 1949.

[Savage, 1954] L. J. Savage. *Foundations of Statistics*. John Wiley and Sons, New York, 1954.

[Shafer, 1976] Glenn Shafer. *A Mathematical Theory of Evidence*. Princeton University Press, Princeton, 1976.

[Smets and Kennes, 1994] Philippe Smets and Robert Kennes. The transferable belief model. *Artificial Intelligence*, 66:191–234, 1994.

[Walley, 1991] Peter Walley. *Statistical Reasoning with Imprecise Probabilities*. Chapman and Hall, London, 1991.

[Zadeh, 1975] Lotfi A. Zadeh. Fuzzy logic and approximate reasoning. *Synthese*, 30, 1975.